\documentclass[twoside]{article}
\usepackage[accepted]{aistats2016}

\usepackage{amsmath,amssymb,amsthm}

\usepackage{microtype}

\usepackage[utf8]{inputenc} 
\usepackage{graphicx} 
\usepackage{subfigure}

\usepackage{algorithm}
\usepackage{algorithmicx}
\usepackage[noend]{algpseudocode} %

\usepackage[square,sort,comma,numbers]{natbib}
\usepackage{afterpage}
\usepackage[colorlinks,citecolor=blue,urlcolor=blue,linkcolor=blue]{hyperref}

\usepackage{wrapfig}
\usepackage{grffile}
\usepackage{float}
\usepackage{rotating}
\usepackage{enumitem} 
\usepackage{bm}
\usepackage{dsfont}
\usepackage{cite}
\usepackage{color}
\usepackage{soul}
\usepackage{wasysym}
\usepackage{url}

\usepackage[capitalize]{cleveref}  
\usepackage{mathtools}
\usepackage{amsthm}
\usepackage{thmtools}

\usepackage[bf,small]{caption} %
\DeclareCaptionType{copyrightbox} %

\newcommand{\real}{\mathrm{I\kern-0.175em R}}

\newcommand{\E}{\mathbb E}

\def\bdelta{\bm{\delta}}

\def\bmu{\bm{\mu}}

\def\btau{\bm{\tau}}

\def\btheta{\bm{\theta}}

\def\D{\mathcal{D}}

\def\O{\mathcal{O}}

\def\bx{x}

\def\bX{X}

\def\Normal{\mathcal{N}}

\newcommand{\Var}{{\rm Var}} 

\def\Reals{\mathbb{R}}
\renewcommand{\bx}{\bm{x}}
\renewcommand{\bX}{\bm{X}}

\newcommand{\TS}{T}%

\newcommand{\defn}[1]{{\bf #1}}

\renewcommand{\min}{\mathsf{min}}
\renewcommand{\max}{\mathsf{max}}

\newcommand{\PT}{\mathsf{T}}

\newcommand{\samplepartialMPblock}{\mathsf{SampleMondrianBlock}}
\newcommand{\samplepartialMPtree}{\mathsf{SampleMondrianTree}}
\newcommand{\extendMPblock}{\mathsf{ExtendMondrianBlock}}
\newcommand{\extendMPtree}{\mathsf{ExtendMondrianTree}}

\newcommand{\predict}{\mathsf{Predict}}
\newcommand{\minsamples}{\mathsf{min\_samples\_split}}
\newcommand{\minsamplesleaf}{\mathsf{min\_samples\_leaf}}
\newcommand{\pnotsplityet}{p_{\mathsf{NotSeparatedYet}}}

\newcommand{\Mondrian}{\mathcal M}

\newcommand{\sdim}{\delta}
\newcommand{\bsdim}{\bdelta}
\newcommand{\sloc}{\xi}
\newcommand{\bsloc}{\bm{\xi}}

\newcommand{\bstime}{\btau}

\newcommand{\parent}{\mathsf{parent}}

\newcommand{\leaf}[1]{\mathsf{leaves}(#1)}
\newcommand{\nonleaf}[1]{#1 \setminus \leaf {#1}} 
\newcommand{\leafx}[1]{\mathsf{leaf}(#1)}

\newcommand{\ancestralpath}{\mathsf{path}}

\newcommand{\nonnegative}[1]{\max(#1, 0)}	
\newcommand{\xdomain}{\Reals^D}

\renewcommand{\bell}{\bm{\ell}}
\newcommand{\bu}{\mathbf{\uu}}
\newcommand{\be}{\mathbf{e}}

\renewcommand{\tt}{\tau}
\newcommand{\uu}{u}

\renewcommand{\root}{\epsilon}

\newcommand{\lleft}{\mathsf{left}}

\newcommand{\mysigmoid}[1]{\gamma_1\sigma(\gamma_2 #1)}

\newcommand{\leftr}{\mathsf{left}}
\newcommand{\leftj}{\mathsf{left}(j)}
\newcommand{\rightj}{\mathsf{right}(j)}
\newcommand{\childj}{\mathsf{child}(j)}

\newcommand{\tj}{\tilde{\jmath}}

\newcommand{\algcomment}[1]{\Comment{\textit{#1}}}
\algnewcommand{\LineComment}[1]{\(\triangleright\) \textit{#1}}

\newcommand{\defas}{:=}

\def\[#1\]{\begin{align}#1\end{align}}

\newcommand{\xtest}{\bx}
\newcommand{\ytest}{y}
\newcommand{\predictive}[1]{p_{#1}(\ytest|\xtest, \D_{1:N})}

\newlength{\figwidth}
\newlength{\sfigwidth}
\crefname{algocf}{alg.}{algs.}
\Crefname{algocf}{Algorithm}{Algorithms}
\crefname{lem}{Lemma}{Lemmas}
\crefname{prop}{Proposition}{Propositions}
\crefname{cor}{Corollary}{Corollaries}
\crefname{thm}{Theorem}{Theorems}
\crefname{assumption}{Assumption}{Assumptions}

\begin{document}

\twocolumn[
\aistatstitle{Mondrian Forests for Large-Scale Regression  \\ when Uncertainty Matters}
\aistatsauthor{ Balaji Lakshminarayanan \And Daniel M. Roy \And Yee Whye Teh }
\aistatsaddress{ Gatsby Unit\\University College London  \And Department of Statistical Sciences\\ University of Toronto \And Department of Statistics\\ University of Oxford } 
]

\begin{abstract} 

Many real-world regression problems demand a measure of the uncertainty associated with each prediction.  
Standard decision forests deliver efficient state-of-the-art predictive performance, but high-quality uncertainty estimates are lacking.
Gaussian processes (GPs) deliver uncertainty estimates, but scaling GPs to large-scale data sets %
 comes at the cost of approximating the uncertainty estimates. 
We extend Mondrian forests, first proposed by Lakshminarayanan et al. (2014) 
for classification problems, to the large-scale non-parametric regression setting.
Using a novel hierarchical Gaussian prior that dovetails with the Mondrian forest framework, we obtain principled uncertainty estimates, %
 while still retaining the computational advantages of decision forests.
Through a combination of illustrative examples, real-world large-scale datasets, and Bayesian optimization benchmarks,
we demonstrate that Mondrian forests %
 outperform approximate GPs on large-scale regression tasks and deliver better-calibrated uncertainty assessments than decision-forest-based methods.

\end{abstract}

\section{Introduction}

Gaussian process (GP) regression is popular due to its ability to deliver both 
accurate non-parametric predictions and estimates of uncertainty for those predictions.
The dominance of GP regression in applications such as Bayesian optimization,  where uncertainty estimates are key to balance exploration and exploitation, is a testament to the quality of GP uncertainty estimates. 

Unfortunately, the computational cost of GPs is cubic in the number of data points, making them computationally very expensive for large scale non-parametric regression tasks. (Specifically, we focus on the scenario where the number of data points $N$ is large, but the number of dimensions $D$ is modest.) %
Steady progress has been made over the past decade on scaling GP inference to big data, including some impressive recent work such as \citep{dgp, distgp, svigp}.

Ensembles of randomized decision trees, also known as \emph{decision forests}, are popular for (non-probabilistic) non-parametric regression tasks, often achieving state-of-the-art  predictive performance \citep{caruana2006empirical}. The most popular decision forest variants are \emph{random forests} (Breiman-RF) introduced by \citet{RF} and \emph{extremely randomized trees} (ERT) introduced by \citet{ERT}. The computational cost of 
{learning } 
 decision forests is typically $\O(N\log N)$ and the computation across the trees in the forest can be parallelized trivially, making them attractive for large scale regression tasks. While decision forests usually yield good predictions (as measured by, e.g., mean squared error or classification accuracy), the uncertainty estimates of decision forests are not as good as those produced by GPs. For instance, \citet{kjit} compare the uncertainty estimates of decision forests and GPs on a simple regression problem where the test distributions are different from the training distribution. As we move away from the training distribution, GP predictions smoothly return to the prior and exhibit higher uncertainty. However, the uncertainty estimates of decision forests are less smooth and do not exhibit this desirable property.

Our goal is to combine the desirable properties of GPs (good uncertainty estimates, probabilistic setup) with those of decision forests (computational speed). To this end, we extend Mondrian forests (MFs), introduced by \citet{MF} for classification tasks, to non-parametric regression tasks. 
Unlike usual decision forests, we use a probabilistic model within each tree to model the labels. Specifically, we use a hierarchical Gaussian prior over the leaf node parameters and compute the posterior parameters efficiently using Gaussian belief propagation \citep{murphy2012machine}. Due to special properties of Mondrian processes, their use as the underlying randomization mechanism results in a desirable uncertainty property:
the prediction at a test point shrinks to the prior as the test point moves further away from the observed training data points. We demonstrate that, as a result, MFs %
 yield better uncertainty estimates. %

The paper is organized as follows: in Section \ref{sec:mf}, we briefly review Mondrian forests. We present MFs %
for regression in Section \ref{sec:label distribution} and discuss inference and prediction in detail. We present experiments in Section \ref{sec:experiments} that demonstrate that (i) MFs %
 produce better uncertainty estimates than Breiman-RF and ERT when test distribution is different from training distribution, (ii) MFs %
  outperform or
achieve comparable performance to large scale approximate GPs  in terms of mean squared error (MSE) or negative log predictive density (NLPD), thus making them well suited for large scale regression tasks where uncertainty estimates are important, and 
 (iii) MFs %
 outperform (or perform as well as) decision forests on Bayesian optimization tasks, where predictive uncertainty is important (since it guides the exploration-exploitation tradeoff). 
Finally, we discuss avenues for future work in Section \ref{sec:discussion}.

\section{Mondrian forests}\label{sec:mf}
\citet{MF} introduced Mondrian forests (MFs) for classification tasks. 
For completeness, we briefly review decision trees and Mondrian trees before describing how MFs can be applied to regression. 
Our problem setup is the following: given $N$ labeled examples $(\bx_1,y_1), \dotsc, (\bx_N,y_N) \in \xdomain \times \Reals$  
as training data, 
our task is to predict labels\footnote{We refer to $y\in\Reals$ as label even though it is common in statistics to refer to $y\in\Reals$ as response instead of label.}
 $y \in \Reals$ for unlabeled test points $\bx \in \xdomain$
as well as provide corresponding estimates of uncertainty.
Let $\bX_{1:n}\defas(\bx_{1},\dotsc,\bx_n)$, $Y_{1:n}\defas(y_1,\dotsc,y_n)$, and $\mathcal D_{1:n}\defas(\bX_{1:n},Y_{1:n})$.

\begin{figure}%
\begin{center}
\includegraphics[height=0.945in]{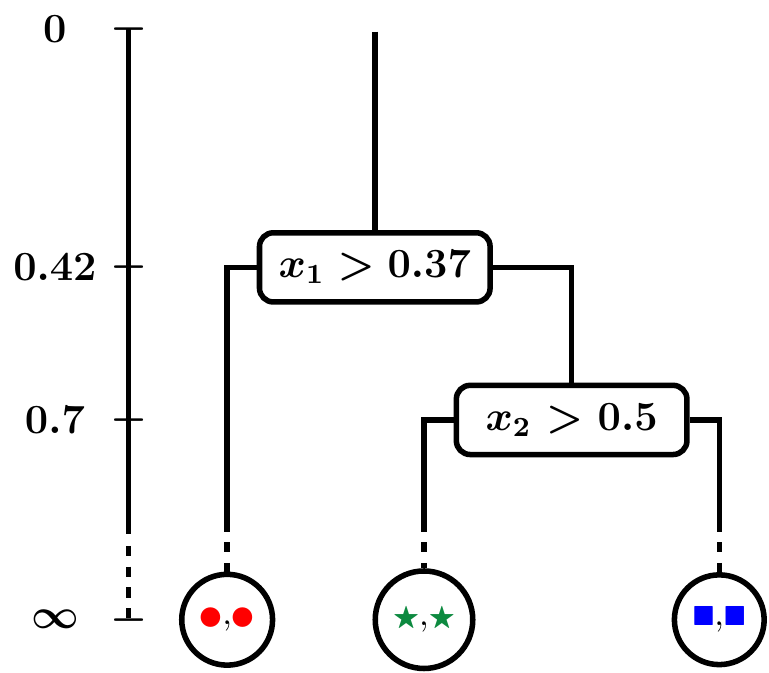}
\includegraphics[height=0.945in]{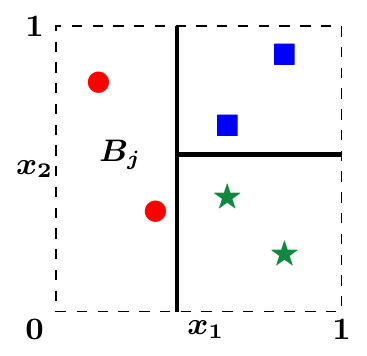}\hspace{1em}
\includegraphics[height=0.945in]{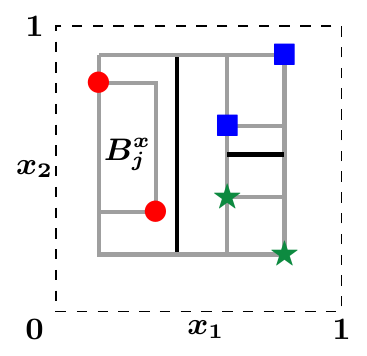}
\end{center}
\vspace{-0.1in}
\caption{%
{\bf (left)} A Mondrian tree over six \emph{data points} in $[0,1]^2$.
Every node in the tree represents a split and is embedded in time (vertical axis).
{\bf (middle)} An ordinary decision tree partitions the whole space.
{\bf (right)} In a Mondrian tree, splits are committed only within the range of the data in each block (denoted by gray rectangles).
Let $j = \leftr(\root)$ be the left child of the root: then $B_j=(0,0.37]\times(0,1]$ is the block containing the red circles and $B_j^x\subseteq B_j$ is the smallest rectangle enclosing the two data points.
(Adapted from \citep{MF}, with permission.)
}
\label{fig:mtreevsdtree}
\end{figure}

\subsection{Decision trees}
Following \citep{MF},
a decision tree is a triple
$(\PT,\bsdim,\bsloc)$
where 
$\PT$ is a finite, rooted, strictly binary tree and
$\bsdim=(\bsdim_j)$ and $\bsloc=(\bsloc_j)$ specify, 
for every internal node $ j \in \nonleaf\PT$, a \defn{split dimension} $\bsdim_j \in \{1,\dotsc,D\}$ 
and \defn{split location} $\bsloc_j \in \Reals$.
A decision tree represents a hierarchical partition of $\Reals^D$ into blocks, one for each node, as follows:
At the root node, $\root$, we have $B_{\root}=\Reals^D$, while each internal node $j\in\nonleaf\PT$ 
represents a \defn{split} of its parent's block into two halves, 
with $\sdim_{j}$ denoting the dimension of the split and $\sloc_{j}$ denoting the location of the split.  
In particular,
\begin{align*} %
 B_{\leftj} &\defas \{\bx\in B_j: x_{\sdim_j}\leq\sloc_j\} \quad \textrm{and} \quad \\
  B_{\rightj} &\defas \{\bx\in B_j: x_{\sdim_j}>\sloc_j\}.
\end{align*}
 We will write
 $B_j=\bigl(\ell_{j1}, \uu_{j1}\bigr]\times\ldots\times\bigl(\ell_{jD}, \uu_{jD}\bigr]$,
where $\ell_{jd}$ and $u_{jd}$ denote %
the $\ell$ower and $u$pper bounds, %
 respectively, 
of the rectangular block $B_j$ along dimension $d$.
Put $\bell_j=\{\ell_{j1}, \ell_{j2}, \ldots, \ell_{jD}\}$ and $\bu_j=\{\uu_{j1}, \uu_{j2}, \ldots, \uu_{jD}\}$.    
Let $N(j)$ denote the indices of training data points at node $j$.

\subsection{Mondrian trees and Mondrian forests}

A Mondrian process \citep{RoyTeh2009a} is a continuous-time Markov process $( \Mondrian_t : t \ge0 )$, where, for every $t \ge s \ge 0$,
$\Mondrian_t$ is a hierarchical binary partition of $\xdomain$ and a refinement of $\Mondrian_s$.
\defn{Mondrian trees} \citep{MF} are restrictions of Mondrian processes to a finite set of points. (See Figure~\ref{fig:mtreevsdtree}.)
In particular, a Mondrian tree $T$ is a tuple $(\PT, \bsdim, \bsloc, \bstime)$, where $(\PT,\bsdim,\bsloc)$ is a decision tree and $\bstime=\{\tt_j\}_{j\in\PT}$ specifies a \defn{split time} $\tt_j \ge 0$ with each node $j$. 
Split times increase with depth, i.e., $\tt_j>\tt_{\parent(j)}$ and play an important role in online updates.

The expected depth of a Mondrian tree is parametrized by a non-negative \emph{lifetime} parameter $\lambda > 0$. Since it is difficult to specify $\lambda$, \citet{MF} set $\lambda=\infty$ and %
stopped splitting a node if all the class labels of the data points within the node were identical. We follow a similar approach for regression: we do not split a node which has less than $\minsamples$ number of data points.\footnote{Specifying $\minsamples$ instead of max-depth is common in decision forests, cf.~\citep{ERT}.}
Given a bound $\minsamples$ and training data $\D_{1:n}$, the generative process for sampling Mondrian trees 
is described in Algorithms~\ref{alg:generative process} and~\ref{alg:generative process:partialblock}. 

\begin{algorithm*}%
\caption{$\samplepartialMPtree\bigl(\D_{1:n}, \minsamples\bigr)$}
\label{alg:generative process}
\begin{algorithmic}[1]
\State Initialize: $\PT=\emptyset$, $\leaf\PT=\emptyset$, $\bsdim=\emptyset$, $\bsloc=\emptyset$, $\bstime=\emptyset$, $N(\root)=\{1,2,\ldots, n\}$ \algcomment{initialize empty tree}
\State $\samplepartialMPblock\bigl(\root, \D_{N(\root)},\minsamples\bigr)$ \algcomment{Algorithm~\ref{alg:generative process:partialblock}}
\end{algorithmic}
\end{algorithm*}
\begin{algorithm*}%
\caption{$\samplepartialMPblock\bigl(j, \D_{N(j)},\minsamples\bigr)$}
\label{alg:generative process:partialblock}
\begin{algorithmic}[1]
\State Add $j$ to $\PT$ and 
for all $d$, set $\ell_{jd}^x=\min(\bX_{N(j),d}), \uu_{jd}^x=\max(\bX_{N(j),d})$ \algcomment{dimension-wise $\min$ and $\max$}
\If{$|N(j)| \geq \minsamples$ } \algcomment{$j$ is an internal node. $|N(j)|$ denotes  $\#$
 data points.}
\State Sample $E$ from exponential distribution with rate $\sum_d (u_{jd}^x-\ell_{jd}^x)$
\State Set $\tt_j = \tt_{\parent(j)}+E$
\State Sample split dimension $\sdim_j$, choosing $d$ with probability proportional to $u_{jd}^x-\ell_{jd}^x$
\State Sample split location $\sloc_j$ uniformly from interval $[\ell_{j\sdim_j}^x, u_{j\sdim_j}^x]$ %
\State Set $N(\leftj)=\{n\in N(j): \bX_{n,\sdim_j}\leq\sloc_j\}$ and $N(\rightj)=\{n\in N(j): \bX_{n,\sdim_j}>\sloc_j\}$
\State $\samplepartialMPblock\bigl(\leftj,  \D_{N(\leftj)}, \minsamples\bigr)$ 
\State $\samplepartialMPblock\bigl(\rightj, \D_{N(\rightj)}, \minsamples\bigr)$ 
\Else \algcomment{$j$ is a leaf node}
\State Set $\tt_j=\infty$ and add $j$ to $\leaf\PT$
\EndIf 
\end{algorithmic}
\end{algorithm*}

The process is similar to top-down induction of decision trees except for the following key differences: (i) splits in a Mondrian tree are committed only within the range of training data (see Figure~\ref{fig:mtreevsdtree}), and (ii) the split dimensions and locations are chosen independent of the labels and uniformly within $B_j^x$ (see lines 5, 6 of Algorithm~\ref{alg:generative process:partialblock}). A \defn{Mondrian forest} consists of $M$ i.i.d.~Mondrian trees $\TS_m=(\PT_m, \bsdim_m, \bsloc_m,\bstime_m)$ for $m=1,\dotsc,M$. 
See \citep{MF} for further details. 

Mondrian trees can be updated online in an efficient manner and remarkably, the distribution of trees sampled from the online algorithm is identical to the corresponding batch counterpart \citep{MF}. We use the batch version of Mondrian forests (Algorithms~\ref{alg:generative process} and \ref{alg:generative process:partialblock}) in all of our experiments except the Bayesian optimization experiment in section~\ref{sec:bayesopt}. Since we do not evaluate the computational advantages of online Mondrian forest, using a batch Mondrian forest in the Bayesian optimization experiment would not affect the reported results.  For completeness, we describe the online updates in Algorithms~\ref{alg:conditionalmondrian} and \ref{alg:extendmondrianblock} in the supplementary material.

\section{Model, hierarchical prior, and predictive posterior for labels}\label{sec:label distribution} 

In this section, %
we describe a probabilistic model that will determine the predictive label distribution, $\predictive {\TS}$, for a tree $\TS = (\PT, \bsdim, \bsloc, \bstime)$, dataset $\D_{1:N}$, and test point~$\bx$.  
Let $\leafx{\bx}$ denote the unique leaf node $j \in \leaf{\PT}$ such that $\bx\in B_{j}$.
Like with Mondrian forests for classification, we want the predictive label distribution at $\bx$ to be a smoothed version of the empirical distribution of labels for points in $B_{\leafx{\bx}}$ and in $B_{j'}$ for nearby nodes $j'$.
We will also achieve this smoothing via a hierarchical Bayesian approach:  every node is associated with a label distribution, and a prior is chosen under which the label distribution of a node is similar to that of its parent's.  The predictive $\predictive {\TS}$ is then obtained via marginalization.

As is common in the decision tree literature, we assume the labels within each block are independent of $\bX$ given the tree structure.
\citet{MF} used a hierarchy of normalized stable processes (HNSP) prior for classification problems. 
 In this paper, we focus on the case of real-valued labels. 
Let $\Normal(\mu,v)$ denote a Gaussian distribution with mean $\mu$ and variance
$v$.
For every $j \in \PT$, let $\mu_j$ be a mean parameter (of a Gaussian distribution over the labels) at node $j$, %
 and let $\bmu=\{\mu_j:j\in \PT\}$.
We assume 
the labels within a leaf node are Gaussian distributed:
\begin{align}\label{eq:likelihood}
y_n|\TS, \bmu \sim \Normal(\mu_{\leafx{\bx_n}}, \sigma_y^2)
\end{align}
where 
$\sigma_y^2$ is a parameter specifying the variance of the (measurement) noise.

We use the following hierarchical Gaussian prior for $\bmu$:
For hyperparameters $\mu_H$, $\gamma_1, \gamma_2$, let
\begin{align*}
\mu_\root|\mu_H \sim \Normal(\mu_H, \phi_\root),
\  \ %
\mu_{j}|\mu_{\parent(j)} \sim \Normal(\mu_{\parent(j)}, \phi_j),
\end{align*}
where
$\phi_j=\mysigmoid{\tt_j} -\mysigmoid{\tt_{\parent(j)}}$ with the convention that $\tt_{\parent(\root)}=0$, 
and $\sigma(t) = (1+e^{-t})^{-1}$ denotes the sigmoid function.

Before discussing the details of posterior inference, we provide some justification for the details of this model:
 Recall that 
 $\tt_j$ %
 increases as we go down the tree, and so the use of the sigmoid $\sigma(\cdot)$ encodes the prior assumption that children are expected to be more similar to their parents as depth increases.  
 The Gaussian hierarchy is \emph{closed under marginalization}, i.e.,
\begin{align*}
\begin{aligned}
 \mu_\root|\mu_H &\sim \Normal(\mu_H,\phi_\root)  
 \\ \mu_0|\mu_\root, \mu_H &\sim \Normal(\mu_\root, \phi_0)
\end{aligned}
\ \Rightarrow \  %
  \mu_0|\mu_H\sim\Normal(\mu_H, \phi_\root+\phi_0), 
\end{align*}
where $\phi_\root+\phi_0=\mysigmoid{\tt_\root}-\mysigmoid{0}+\mysigmoid{\tt_0}-\mysigmoid{\tt_\root}=\mysigmoid{\tt_0}-\mysigmoid{0}$. 
Therefore, we can introduce intermediate nodes without changing the predictive distribution. In Section~\ref{sec:prediction}, we show that a test data point can branch off into its own node: the hierarchical prior is critical for smoothing predictions.

Given training data $\D_{1:N}$, %
our goal is to compute the posterior density over $\bmu$:
\begin{align}
p_{\TS}(\bmu|\D_{1:N})\propto p_{\TS}(\bmu) \prod_{n=1}^N \Normal(y_n|\mu_{\leafx{\bx_n}},\sigma_y^2).
\end{align} 
The posterior over $\bmu$ will be used during the prediction step described in Section~\ref{sec:prediction}.
Note that the posterior over $\bmu$ is computed independently for each tree, and so can be parallelized trivially.

\subsection{Gaussian belief propagation}

We perform posterior inference using belief propagation \citep{pearl1988probabilistic}. 
Since the prior and likelihood are Gaussian, all the messages can be computed analytically and the posterior over $\bmu$ is also Gaussian. 
Since the hierarchy has a tree structure, the posterior can be computed in time that scales linearly with the number of nodes in the tree, which is typically $\O(N)$, hence posterior inference is efficient compared to non-tree-structured Gaussian processes whose computational cost is typically $\O(N^3)$.
Message passing in trees is a well-studied problem, and so we refer the reader to \citep[][Chapter~20]{murphy2012machine} for details.

\subsection{Hyperparameter heuristic}\label{sec:hyper}

In this section, we briefly describe how we choose the hyperparameters $\btheta=\{\mu_H, \gamma_1, \gamma_2, \sigma_y^2\}$.  More details can be found in 
Appendix~\ref{app:hyper} in the supplementary material. 
For simplicity, we use the same values of these hyperparameters for all the trees; 
it is possible to optimize these parameters for each tree independently, and would be interesting to evaluate this extra flexibility empirically.
Ideally, one might choose hyperparameters by optimizing the marginal likelihood (computed as a byproduct of belief propagation) by, e.g., gradient descent.  We use a simpler approach here: we maximize the product of the individual label marginals, assuming a individual label noise, which yields closed-form solutions for $\mu_H$ and $\gamma_1$. %
  This approach does not yield an estimate for $\gamma_2$,
and so, using the fact that $\tt$ increases with $N$, we pre-process the training data to lie in $[0,1]^D$ and set $\gamma_2= D / (20\log_2 N)$ based on the reasoning that (i) $\tt$ is inversely proportional to $D$ and (ii) $\tt$ increases with tree depth and the tree depth is $\O(\log_2 N)$ assuming balanced trees.\footnote{In \citet{MF}, it was observed that the average tree depths were 2-3 times $\log_2(N)$ in practice.}

\citet{MF} proposed to stop splitting a Mondrian block whenever all the class labels were identical.\footnote{Technically, the Mondrian tree is \emph{paused} in the online setting and splitting resumes once a block contains more than one distinct label.  However, since we only deal with the batch setting, we stop splitting homogeneous blocks.} 
We adopt a similar strategy here and stop splitting a Mondrian block if the number of samples is fewer than a parameter $\minsamples$. It is common in decision forests to require a minimum number of samples in each leaf, for instance \citet{RF} and \citet{ERT} recommend setting $\minsamplesleaf=5$ for regression problems. 
We set $\minsamples=10$.

\subsection{Predictive variance computation}\label{sec:prediction}

The prediction step in a Mondrian regression tree is similar to that in a Mondrian classification tree \citep[Appendix B]{MF} except that at each node of the tree, we predict a Gaussian posterior over $y$ rather than predict a posterior over class probabilities. 
Recall that a prediction from a vanilla decision tree is just the average of the training labels in $\leafx{\bx}$. 
Unlike decision trees, in a Mondrian tree, a test point $\bx$ \emph{can potentially `branch off' the existing Mondrian tree at any point along the path from root to $\leafx{\bx}$}.  Hence, the predictive posterior over $y$ from a given tree $\TS$ is a mixture of Gaussians of the form 
\begin{align}
\predictive{\TS}=& 
\sum_{j\in\ancestralpath(\leafx{\bx})} w_j \Normal(y|m_j, v_j),
\label{eq:pygivenx_tree}
\end{align}
where $w_j$ denotes the weight of each component, given by the probability of branching off just before reaching node $j$, and $m_j,v_j$ respectively denote the predictive mean and variance. %
The probability of branching off increases as the test point moves further away from the training data at that particular node;  hence, the predictions of MFs exhibit higher uncertainty as we move farther from the training data.  
For completeness, we provide pseudocode for computing (\ref{eq:pygivenx_tree}) in Algorithm~\ref{alg:predict} of the supplementary material.%

If a test data point branches off to create a new node, the predictive mean at that node is the posterior of the parent of the new node; if we did not have a hierarchy and instead assumed the predictions at leaves were i.i.d, then branching would result in a prediction from the prior, which would lead to biased predictions in most applications.
The predictive mean and variance for the mixture of Gaussians are 
\[ %
\E_{\TS}[y] &= \sum_j w_j m_j 
 \qquad\text{and}\qquad \nonumber \\
 \Var_{\TS}[y] &= \sum_j w_j (v_j + m_j^2) - \bigl(\E_{\TS}[y]\bigr)^2,
\label{eq:mog}
\]
and the prediction of the ensemble is then
\begin{align}{\label{eq:predensemble}
p(y|\bx,\D_{1:N}) = \frac{1}{M} \sum_m \predictive{\TS_m}.
}\end{align}
The prediction of the ensemble can be thought of as being drawn from a mixture model over $M$ trees where the trees are weighted equally. %
The predictive mean and variance of the ensemble can be computed using the formula for mixture of Gaussians similar to (\ref{eq:mog}). 
Similar strategy has been used in \citep{DF,hutter2014algorithm} as well. %

\section{Related work}\label{sec:related work}

 The work on large scale Gaussian processes can be broadly split into approaches that optimize inducing variables using variational approximations and approaches that distribute computation by using experts that operate on subsets of the data. We refer to \citep{dgp} for a recent summary of large scale Gaussian processes. 
 \citet{svigp} and  \citet{distgp} use stochastic variational inference to speed up GPs, building on the variational bound developed by \citet{titsias2009variational}.
 \citet{dgp} present the robust Bayesian committee machine (rBCM), which combines predictions from experts that operate on subsets of data.

\citet{hutter2009automated} investigated the use of Breiman-RF for Bayesian optimization and used the empirical variance between trees in the forest as a measure of uncertainty. (\citet{hutter2014algorithm} proposed a further modification, see Appendix~\ref{app:fast}.)
\citet{eslami} used a non-standard decision forest implementation where a quadratic regressor is fit at each leaf node, rather than a constant regressor as in popular decision forest implementations.
Their uncertainty measure---a sum of the Kullback-Leibler (KL) divergence---is highly specific to their application of accelerating expectation propagation,
and so it seems their method is unlikely to be immediately applicable to general non-parametric regression tasks. 
Indeed, \citet{kjit} demonstrate that the uncertainty estimates proposed by  \citep{eslami} are not as good as kernel methods in their application domain when the test distribution is different from the training distribution. 
As originally defined, Mondrian forests produce uncertainty estimates for categorical labels, 
but \citet{MF} evaluated their performance on (online) prediction (classification accuracy) without any assessment of the uncertainty estimates.

\section{Experiments}\label{sec:experiments}

\subsection{Comparison of uncertainty estimates of MFs to popular decision forests}
In this experiment, we compare uncertainty estimates of MFs to those of popular decision forests. The prediction of MFs is given by (\ref{eq:predensemble}), from which we can compute the predictive mean and predictive variance.\footnote{Code available from the authors' websites.} 
For decision forests, we compute the predictive mean as the average of the predictions from the individual trees and, 
following \citet[\S11.1.3]{hutter2009automated}, 
compute the predictive variance as the variance of the predictions from the individual trees. We use 25 trees and set $\minsamplesleaf=5$ for decision forests to make them comparable to MFs with $\minsamples=10$. We used the ERT and Breiman-RF implementation in \emph{scikit-learn} \citep{scikit-learn} and set the remaining hyperparameters to their default values.

\begin{figure*}[htbp] %
\centering
  \subfigure[Distribution of train/test inputs (labels not depicted)]{
   \includegraphics[width=\figwidth]{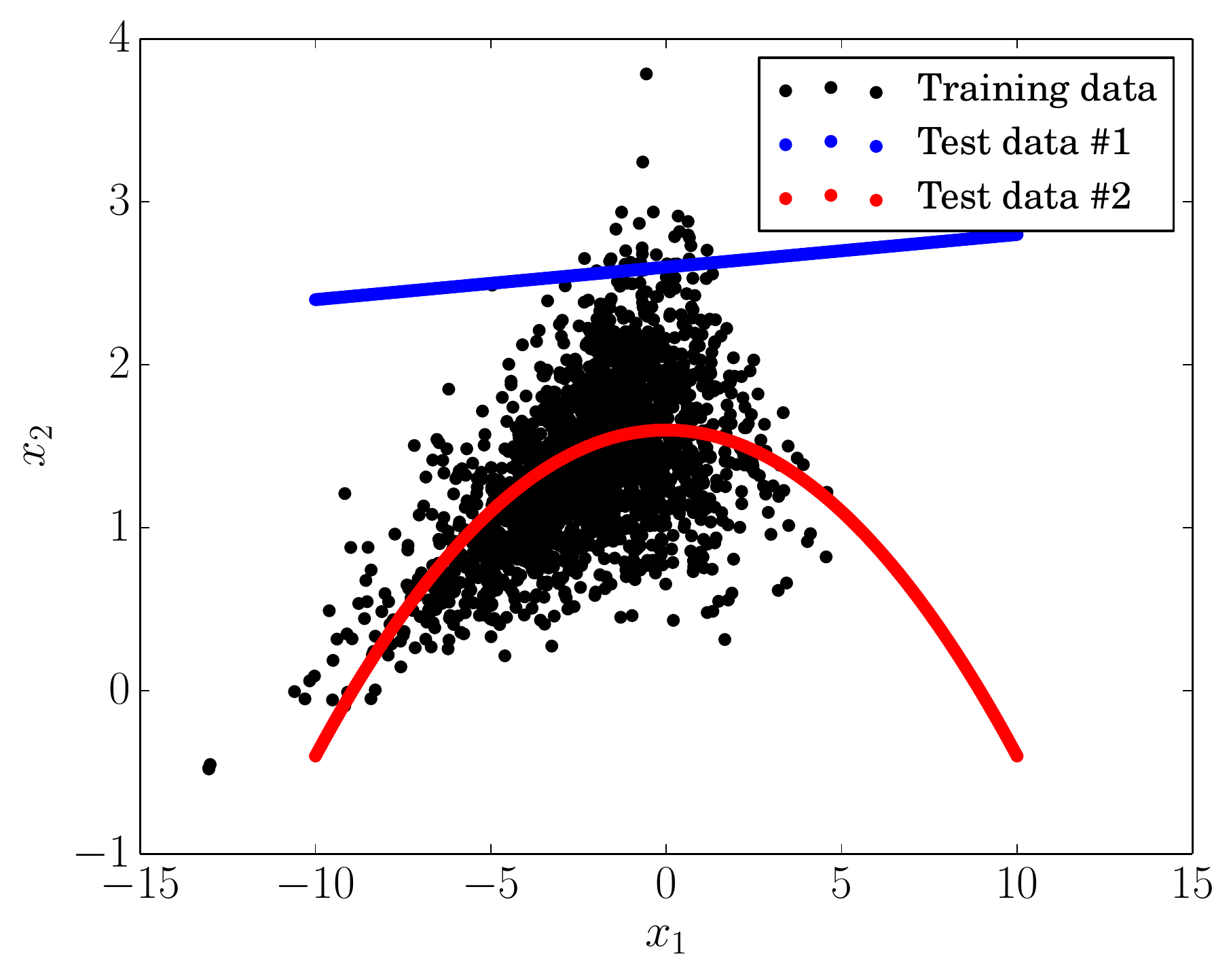}
 \label{fig:data} }
      \subfigure[MF uncertainty]{
   \includegraphics[width=\figwidth]{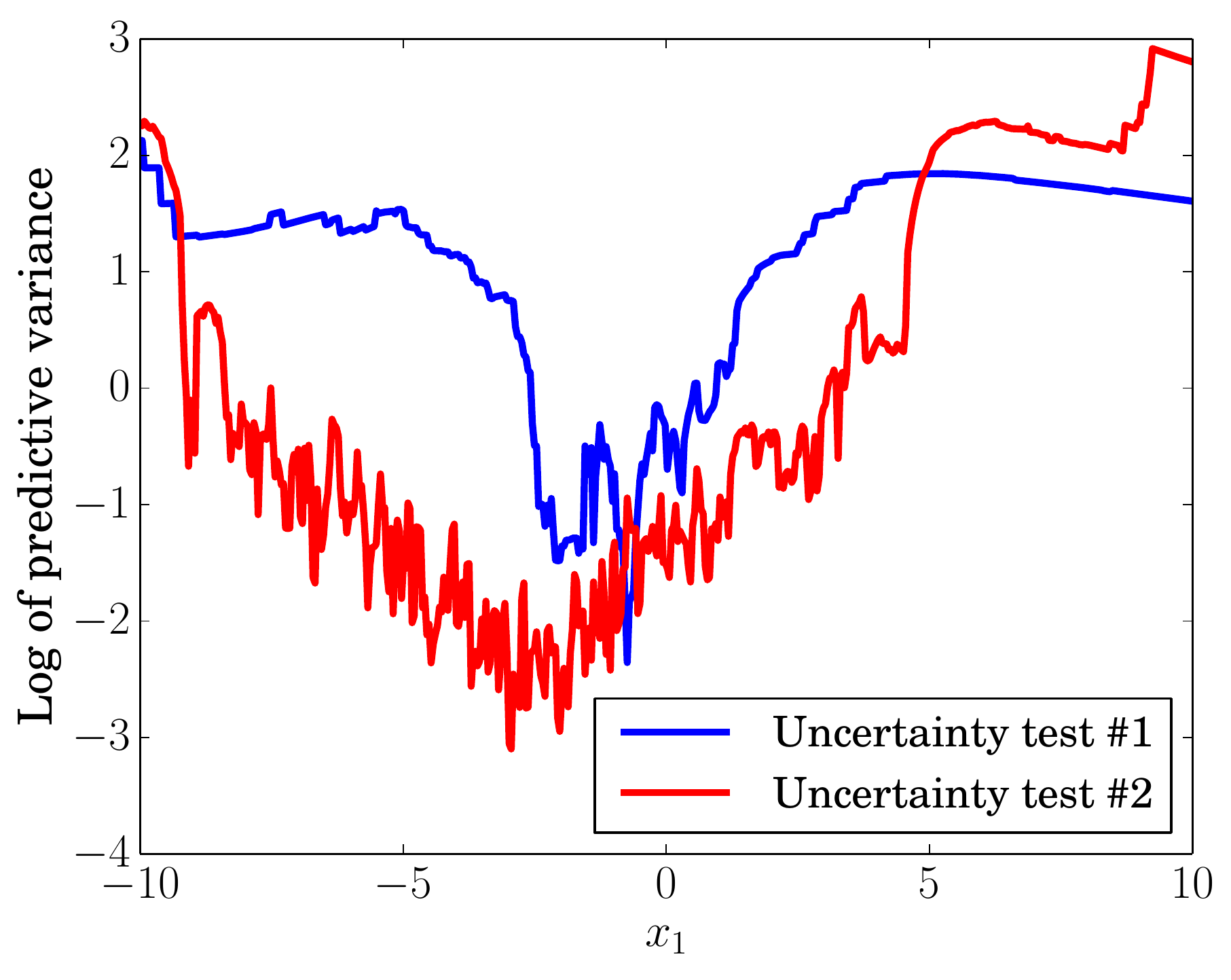}
    \label{fig:uncertainty:mf}
  }
       \subfigure[ERT uncertainty]{
   \includegraphics[width=\figwidth]{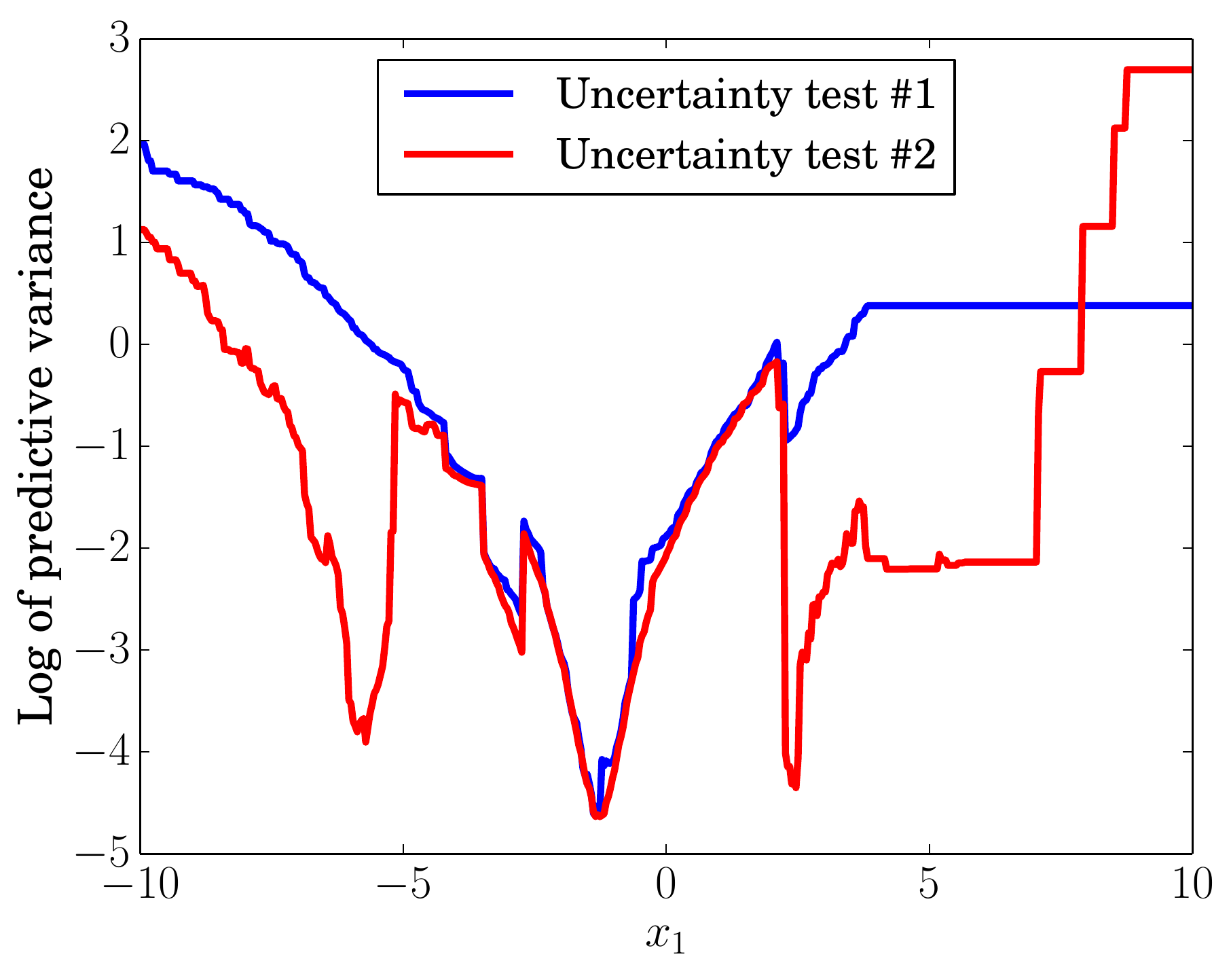}
   \label{fig:uncertainty:ert}
   }
   \subfigure[Breiman-RF uncertainty]{
   \includegraphics[width=\figwidth]{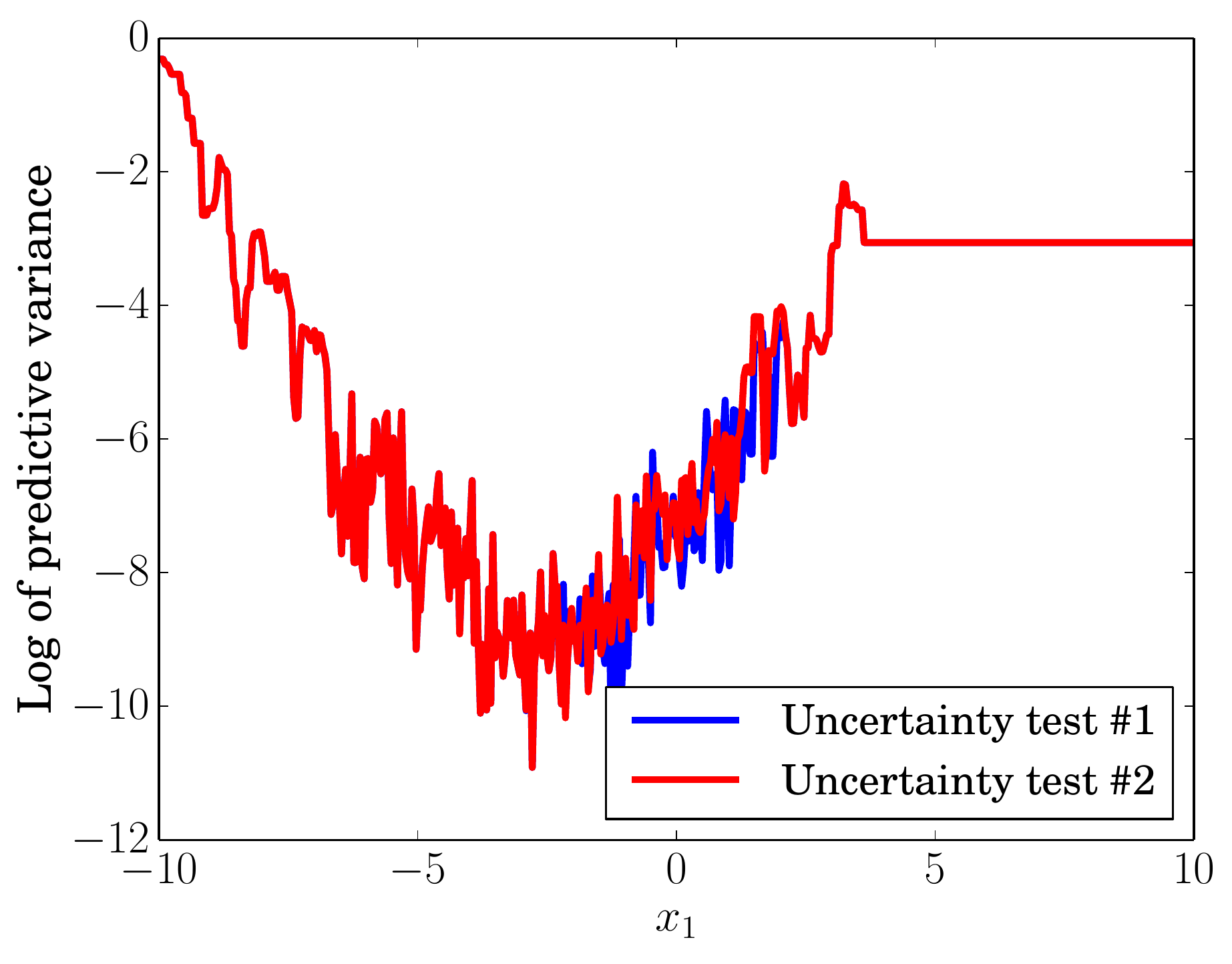}
  \label{fig:uncertainty:rf}
   }
  \caption{ (a) Scatter distribution of training distribution and test distributions. %
  (b-d) Typical uncertainty estimates from a single run of MF, ERT-$k$ and Breiman-RF, respectively, as a function of $x_1$. (Averaging over multiple runs would create smooth curves while obscuring interesting patterns in %
   the estimates which an application would potentially suffer from.)  As desired, MF becomes less certain away from training inputs, while the other methods report high confidence spuriously.
  }
  \label{fig:logistic_uncertainty}
\end{figure*}

We use a simplified version of the dataset described in \citep{kjit}, where the goal is to predict the outgoing message in expectation propagation (EP) from a set of incoming messages. When the predictions are uncertain, the outgoing message will be re-computed (either numerically or using a sampler), hence predictive uncertainty is crucial in this application. Our dataset consists of two-dimensional features (which are derived from the incoming message) and a single target (corresponding to mean of the outgoing message). The scatter plot of  the training data features is shown in Fig.~\ref{fig:data}. We evaluate predictive uncertainty on two test distributions, shown in red and blue in  Fig.~\ref{fig:data}, which contain data points in unobserved regions of the training data. 

The mean squared error of all the methods are comparable, so we focus just on the predictive uncertainty.  Figures~\ref{fig:uncertainty:mf}, \ref{fig:uncertainty:ert}, and \ref{fig:uncertainty:rf} display the predictive uncertainty of  MF, ERT and Breiman-RF as a function of $x_1$. We notice that Breiman-RF's predictions are over-confident compared to MF and ERT. The predictive variance is quite low even in regions where training data has not been observed. The predictive variance of MF is low in regions where training data has been observed ($-5<x_1<5$) and goes up smoothly as we move farther away from the training data; the red test dataset is more similar to the training data than the blue test data and the predictive uncertainty of MF on the blue dataset is higher than that of the red dataset, as one would expect. ERT is less overconfident than Breiman-RF, however its predictive uncertainty is less smooth compared to that of MF. 

\subsection{Comparison to GPs and decision forests on flight delay dataset}
In this experiment, we compare decision forest variants to large scale Gaussian processes. 
\citet{dgp} evaluated %
a variety of large scale Gaussian processes on the \emph{flight delay} dataset, %
 processed by \citet{svigp}, and demonstrate that their method achieves state-of-the-art predictive performance; we evaluate decision forests on the same dataset so that our predictive performance can be  directly compared to large scale GPs. 
The goal is to predict the flight delay from eight attributes, namely, the age of the aircraft (number of years since deployment), distance that needs to be covered, airtime, departure time, arrival time, day of the week, day of the month and month.

\citet{dgp} employed the following strategy: train using the first $N$ data points and use the following 100,000 as test data points. \citet{dgp} created three datasets, setting $N$ to $700K$, $2M$ (million) and $5M$ respectively. We use the same data splits and train MF, Breiman-RF, ERT on these datasets so that our results are directly comparable.\footnote{ \citet{distgp} report the performance of Breiman-RF on these datasets, but they restricted the maximum depth of the trees to 2, which hurts the performance of Breiman-RF significantly. They also use a random train/test split, hence our results are not directly comparable to theirs due to the non-stationarity in the dataset.}
 We used 10 trees for each forest to reduce the computational burden. 
 
 We evaluate performance by measuring the root mean squared error (RMSE) and negative log predictive density (NLPD).  NLPD, defined as the negative logarithm of 
\eqref{eq:predensemble}, is a popular measure for measuring predictive uncertainty (cf.~\citep[section 4.2]{quinonero2006evaluating}). NLPD penalizes over-confident as well as under-confident predictions since it not only accounts for predictive mean but also the predictive variance. RF and ERT do not offer a principled way to compute NLPD. But, as a simple approximation, we computed NLPD for RF and ERT assuming a Gaussian distribution with mean equal to the average of trees' predictions, variance equal to the variance of trees' predictions. 

Table~\ref{tab:gp} presents the results. The RMSE and NLPD results for SVI-GP, Dist-VGP and rBCM results were taken from \citep{dgp}, who report a standard error lower than 0.3 for all of their results. Table~1 in \citep{dgp} shows that rBCM achieves significantly better performance than other large scale GP approximations; hence we only report the performance of rBCM here. 
It is important to note that the dataset exhibits non-stationarity: as a result, the performance of decision forests as well as GPs is worse on the larger datasets. (This phenomenon was observed by \citet{distgp} and \citet{dgp} as well.) 
On the $700K$ and $2M$ dataset, we observe that decision forests achieve significantly lower RMSE than rBCM. MF achieves significantly lower NLPD compared to rBCM, which suggests that its uncertainty estimates are useful for large scale regression tasks. However, all the decision forests, including MFs, achieve poorer RMSE performance than rBCM on the larger $5M$ dataset. 
We believe that this is due to the non-stationary nature of the data. 
To test this hypothesis, we shuffled the 5,100,000 data points to create three new training (test) data sets with $5M$ ($100K$) points; all the decision forests achieved a RMSE in the range 31-34 on these shuffled datasets. %

\begin{table*}%
\resizebox{\textwidth}{!}{
\begin{tabular}{c|c c|c c|c c}
& 700K/100K & & 2M/100K & & 5M/100K  & \\
 & RMSE                  & NLPD & RMSE & NLPD & RMSE & NLPD \\ \hline                        
SVI-GP \citep{svigp}                 & 33.0                  & -    & -    & -    & -    & -    \\ 
Dist-VGP \citep{distgp}                 & 33.0                  & -    & -    & -    & -    & -    \\ 
rBCM  \citep{dgp}                  & 27.1                  & 9.1  & 34.4 & 8.4  & \textbf{35.5} & 8.8  \\
RF 		& \textbf{24.07 $\pm$ 0.02} &  5.06 $\pm$ 0.02*    & \textbf{27.3 $\pm$ 0.01}     & 5.19 $\pm$ 0.02*     &    39.47 $\pm$ 0.02  &   \textbf{6.90 $\pm$ 0.05*}   \\ 
ERT                  &        24.32 $\pm$ 0.02               &  6.22 $\pm$ 0.03*    &  27.95 $\pm$ 0.02    & 6.16 $\pm$ 0.01*      &  38.38 $\pm$ 0.02    &  8.41 $\pm$ 0.09*    \\
MF                  &      26.57 $\pm$ 0.04               &  \textbf{4.89 $\pm$ 0.02}    &   29.46 $\pm$ 0.02   & \textbf{4.97 $\pm$ 0.01}     &  40.13 $\pm$ 0.05    & \textbf{6.91 $\pm$ 0.06}  \\  %
\end{tabular}
}
\caption{Comparison of MFs to popular decision forests and large scale GPs on the flight delay dataset. We report average results over 3 runs (with random initializations), along with standard errors. MF achieves significantly better NLPD than rBCM. RF and ERT do not offer a principled way to compute NLPD, hence they are marked with an asterix. 
 }
\label{tab:gp}
\end{table*} %

\begin{table*}%
\center
{
\begin{tabular}{c|c |c c c c c c c c c c|}
Dataset & Method & 0.1 & 0.2 & 0.3 & 0.4 & 0.5 & 0.6 & 0.7 & 0.8 & 0.9 \\
\hline
700K & Breiman-RF & -0.02 & -0.04 &  -0.05 &  -0.06 &  -0.06 &  -0.06 &  -0.05 &  -0.06 &  -0.07 \\
700K & ERT & -0.04 & -0.07 &  -0.11 &  -0.14 &  -0.16 &  -0.18 &  -0.19 &  -0.19 &  -0.18 \\
700K & MF & -0.01 &  -0.02 &  -0.01 &  0 &  0.02 &  0.03 &  0.03 &  0.02 &  0 \\ \hline
2M & Breiman-RF & -0.02 &  -0.04 &  -0.05 &  -0.06 &  -0.05 &  -0.04 &  -0.03 &  -0.03 &  -0.04 \\ 
2M & ERT & -0.04 &  -0.08 &  -0.12 &  -0.15 &  -0.17 &  -0.18 &  -0.19 &  -0.18 &  -0.16 \\
2M & MF & -0.02 &  -0.04 &  -0.05 &  -0.05 &  -0.03 &  0 &  0.02 &  0.03 &  0.01 \\ \hline
5M & Breiman-RF & -0.03 &  -0.06 &  -0.08 &  -0.09 &  -0.1 &  -0.1 &  -0.11 &  -0.1 &  -0.1 \\
5M & ERT & -0.04 &  -0.07 &  -0.11 &  -0.14 &  -0.16 &  -0.18 &  -0.19 &  -0.19 &  -0.18 \\
5M & MF & -0.02 &  -0.04 &  -0.05 &  -0.06 &  -0.06 &  -0.05 &  -0.05 &  -0.05 &  -0.07
\end{tabular}
}
\caption{Comparison of MFs to popular decision forests on the flight delay dataset. Each entry denotes the difference between the observed fraction  minus the ideal fraction (which is shown at the top of the column). Hence, a value of zero implies perfect calibration, a negative value implies overconfidence and a positive value implies under-confident predictor. MF is better calibrated than Breiman-RF and ERT, which are consistently over-confident.}
\label{tab:calibration}
\end{table*} %

MF outperforms rBCM in terms of NLPD on all three datasets. On the 5M dataset, the NLPD of Breiman-RF is similar to that of MF, however Breiman-RF's uncertainty is not computed in a principled fashion. 
As an additional measure of uncertainty, we report \emph{probability calibration measures} (akin to those for binary classification cf.~\url{http://scikit-learn.org/stable/modules/calibration.html}), also known as reliability diagrams \citep{degroot1983comparison}, for MF, Breiman-RF and ERT. %
First, we compute the  $z\%$ (e.g. $90\%$) prediction interval for each test data point based on Gaussian quantiles using predictive mean and variance. Next, we measure what fraction of test observations fall within this prediction interval. For a well-calibrated regressor, the observed fraction should be close to $z\%$. We compute observed fraction for $z=10\%$ to $z=90\%$ in increments of 10. We report observed fraction minus ideal fraction since it is easier to interpret---a value of zero implies perfect calibration, a negative value implies over-confidence (a lot more observations lie outside the prediction interval than expected) and a positive value implies under-confidence. The results are shown in Table~\ref{tab:calibration}.
MF is clearly better calibrated than Breiman-RF and ERT, which seem to be consistently over-confident. Since 5M dataset exhibits non-stationarity, MF appears to be over-confident but still outperforms RF and ERT. \citet{dgp} do not report calibration measures and their code is not available publicly, hence we do not report calibration measures for GPs.

\subsection{Scalable Bayesian optimization}\label{sec:bayesopt}
Next, we showcase the usefulness of MFs in a Bayesian optimization (BayesOpt) task. We briefly review the Bayesian optimization setup for completeness and refer the interested reader to \citep{brochu2010tutorial,snoek2012practical} for further details.  
Bayesian optimization deals with the problem of identifying the global maximizer (or minimizer) of an unknown (a.k.a.~black-box) objective function which is computationally expensive to evaluate.\footnote{For a concrete example, consider the task of optimizing the hyperparameters of a deep neural network to maximize validation accuracy.}  
Our goal is to identify the maximizer in as few evaluations as possible. Bayesian optimization is a model-based sequential search approach to solve this problem. Specifically, given $n$ noisy observations, we fit a \emph{surrogate model} such as a Gaussian process or a decision forest and choose the next location based on an \emph{acquisition function} such as upper confidence bound (UCB) \citep{srinivas2009gaussian} or expected improvement (EI) \citep{mockus1978application}. The acquisition function trades off exploration versus exploitation by choosing input locations where the predictive mean from the surrogate model is high (exploitation) or the predictive uncertainty of the surrogate model is high (exploration). Hence, a surrogate model with well-calibrated predictive uncertainty is highly desirable. 
 Moreover, the surrogate model has to be re-fit after every new observation is added; while this is not a significant limitation for a few  (e.g.~50) observations and scenarios where the evaluation of the objective function is significantly more expensive than re-fitting the surrogate model, the re-fitting can be computationally expensive if one is interested in 
  scalable Bayesian optimization \citep{scalablebayesopt}. 
  
   \citet{hutter2009automated} proposed sequential model-based algorithm configuration (SMAC), which uses Breiman-RF as the surrogate model and the uncertainty between the trees as a heuristic measure of uncertainty.\footnote{\citet[\S4.3.2]{hutter2014algorithm} proposed a further modification to the variance estimation procedure, where each tree outputs a predictive mean and variance, in the spirit of \emph{quantile regression forests} \citep{meinshausen2006quantile}. See Appendix~\ref{app:fast} for a discussion on how this relates to MFs. 
   }
    \citet{nickson2014automated} discuss a scenario where this heuristic produces misleading uncertainty estimates that hinders exploration. It is worth noting that SMAC uses EI as the acquisition function only $50\%$ of the time and uses random search the remaining $50\%$ of the time (which is likely due to the fact that the heuristic predictive uncertainty can collapse to 0). Moreover, SMAC re-fits the surrogate model by running a batch algorithm; the computational complexity of running the batch version $N$ times is $\sum_{n=1}^N \O(n\log n)=\O(N^2\log N)$ \citep{MF}.

MFs are desirable for such an application since they can produce principled uncertainty estimates and can be efficiently updated online with computational complexity $\sum_{n=1}^N \O(\log n)=\O(N\log N)$. %
 Note that the cost of updating the Mondrian tree structure is $\O(\log n)$, however exact message passing costs $\O(n)$. To maintain the $\O(\log n)$ cost, we approximate the Gaussian posterior at each node by a Gaussian distribution whose mean and variance are given by the  empirical mean and variance of the data points at that node. Adding a new data point involves just updating mean and variance for all the nodes along the path from root to a leaf, hence the overall cost is $\O(\log n)$. (See Appendix~\ref{app:fast}.)

We report results on four Bayesian optimization benchmarks used in \citep{eggensperger2013towards,scalablebayesopt}, consisting of two synthetic functions namely the Branin and Hartmann functions, and two real-world problems, namely optimizing the hyperparameters of online latent Dirichlet allocation (LDA) and structured support vector machine (SVM).  LDA and SVM datasets consist of 288 and 1400 grid points respectively; we sampled Branin and Hartmann functions at 250,000 grid points (to avoid implementing a separate optimizer for optimizing over the acquisition function). For SVM and LDA, some dimensions of the grid vary on a non-linear scale (e.g.~$10^{0}, 10^{-1}, 10^{-2}$); we log-transformed such dimensions and scaled all dimensions to $[0,1]$ so that all features are on the same scale. We used 10 trees, set $\minsamples=2$ and use UCB as the acquisition function\footnote{Specifically, we set acquisition function = predictive mean + predictive standard deviation.} for MFs. We repeat our results 15 times (5 times each with 3 different random grids for Branin and Hartmann) and report mean and standard deviation.

 Following \citet{eggensperger2013towards}, we evaluate a fixed number of evaluations for each benchmark and measure the maximum value observed. The results are shown in Table~\ref{tab:bayesopt}. The SMAC results (using Breiman-RF) were taken from Table 2 of \citep{eggensperger2013towards}. Both MF and SMAC identify the optimum for LDA-grid. SMAC does not identify the optimum for Branin and Hartmann functions. We observe that MF finds maxima very close to the true maximum on the grid, thereby suggesting that better uncertainty estimates are useful for better exploration-exploitation tradeoff.  The computational advantage of MFs might not be significant with few evaluations, but we expect MFs to be computationally advantageous in applications with thousands of observations, e.g., scalable Bayesian optimization \citep{scalablebayesopt} and reinforcement learning \citep{ernst2005tree}.

\begin{table}%
\center
\resizebox{1.02\columnwidth}{!}{ %
\begin{tabular}{c|c|c|c}
Dataset (D, \#\textsf{evals}) & Oracle & MF & SMAC \citep{eggensperger2013towards} %
\\ \hline
Branin  (2, 200) & -0.398 & -0.400 $\pm$ 0.005 & -0.655 $\pm$ 0.27 \\
Hartmann (6, 200) & 3.322 & 3.247 $\pm$ 0.109 & 2.977 $\pm$ 0.11 \\
SVM-grid (3, 100) & -1266.2 & -1266.36 $\pm$ 0.52 & -1269.6 $\pm$ 2.9 \\
LDA-grid (3, 50) & -24.1 & -24.1 $\pm$ 0 & -24.1$\pm$ 0.1\\
\end{tabular}
}
\caption{Results on BayesOpt benchmarks: Oracle reports the maximum value on the grid. MF, RF report the maximum value obtained by the respective methods.}

\label{tab:bayesopt}
\end{table} %

\subsection{Failure modes of our approach}
No method is a panacea: here we discuss two failure modes of our approach that would be important to address in future work.
First, we expect GPs to perform better than decision forests on extrapolation tasks; %
 a GP with an appropriate kernel (and well-estimated hyperparameter values) can extrapolate beyond the observed range of training data; however, the predictions of decision forests with \emph{constant} predictors at leaf nodes are confined to the range of minimum and maximum observed $y$. If extrapolation is desired, we need complex regressors (that are capable of extrapolation) at leaf nodes of the decision forest. However, this will increase the cost of posterior inference.  
Second, MFs %
 choose splits independent of the labels; hence irrelevant features can hurt predictive performance \citep{MF}; in the batch setting, one can apply feature selection to filter or down weight the irrelevant features. 

\section{Discussion}\label{sec:discussion}
We developed a novel and scalable methodology for regression based on Mondrian forests that provides both good predictive accuracy as well as sensible estimates of uncertainty.  These uncertainty estimates are important in a range of application areas including probabilistic numerics, Bayesian optimization and planning.  Using a large-scale regression application on flight delay data, we demonstrate that our proposed regression framework can provide both state-of-the-art RMSE and estimates of uncertainty as compared to recent scalable GP approximations. We demonstrate that Mondrian forests deliver better-calibrated uncertainty estimates than existing decision forests, especially in regions far away from the training data. Since Mondrian forests deliver good uncertainty estimates and can be trained online efficiently, they seem promising for applications such as Bayesian optimization and reinforcement learning. %

\section*{Acknowledgments} 
 We thank Wittawat Jitkrittum for sharing the dataset used in \citep{kjit} and helpful discussions. We thank Katharina Eggensperger, Frank Hutter and Ziyu Wang for helpful discussions on Bayesian optimization.   
BL gratefully acknowledges generous funding from the Gatsby Charitable Foundation. 
This research was carried out in part while DMR held a Research Fellowship at Emmanuel College, Cambridge, with funding also from a Newton International Fellowship through the Royal Society. 
YWT's research leading to these results has received funding from EPSRC (grant
EP/K009362/1) and the ERC under the EU's FP7 Programme (grant agreement no.~617411).

\bibliography{mfregression}
\bibliographystyle{abbrvnat} %

\newpage %
\appendix
\onecolumn %
{ \textbf{\Large{Mondrian Forests for Large-Scale Regression when Uncertainty Matters: Supplementary material}}} %

\section{Pseudocode for online learning and prediction}
The online updates are shown in Algorithms~\ref{alg:conditionalmondrian} and \ref{alg:extendmondrianblock}. The prediction step is detailed in Algorithm~\ref{alg:predict}.
\begin{algorithm*}%
\caption{$\extendMPtree(\TS,\D,\minsamples)$} \label{alg:conditionalmondrian}
\begin{algorithmic}[1]
\State Input: Tree $\TS=(\PT, \bsdim, \bsloc, \bstime)$, new training instance $\D=(\bx, y)$ 
\State $\extendMPblock(\TS,\root,\D,\minsamples)$ \algcomment{Algorithm~\ref{alg:extendmondrianblock}}
\end{algorithmic}
\end{algorithm*}
\kern-16pt %
\begin{algorithm*}%
\caption{$\extendMPblock(\TS,j,\D,\minsamples)$} \label{alg:extendmondrianblock}
\begin{algorithmic}[1]
\State Set $\be^\ell = \nonnegative{\bell_j^x - \bx}$ and  $\be^u = \nonnegative{\bx - \bu_j^x}$ \algcomment{$\be^\ell=\be^u=\bm{0}_D$ if $\bx\in B_j^x$}
\State Sample $E$ from exponential distribution with rate $\sum_d (e^\ell_d+e^u_d)$
\If{$\tt_{\parent(j)} + E < \tt_j$}  \algcomment{introduce new parent for node $j$}%
\State Sample split dimension $\sdim$, choosing $d$ with probability proportional to $e^\ell_d+e^u_d$
\State Sample split location $\sloc$ uniformly from interval $[u_{j,\sdim}^x,x_{\sdim}]$ \textbf{if} $x_{\sdim} > u_{j,\sdim}^x$ 
            \textbf{else} $[x_{\sdim},\ell_{j,\sdim}^x]$.

\State Insert a new node $\tj$ just above node $j$ in the tree, and a new leaf $j''$, sibling to $j$, where
\State \qquad $\sdim_{\tj} = \sdim$, $\sloc_{\tj} = \sloc$, 
                       $\tt_{\tj} = \tt_{\parent(j)} + E$, 
                      $\bell_{\tj}^x = \min(\bell_j^x, \bx)$, $\bu_{\tj}^x=\max(\bu_j^x, \bx)$
\State \qquad $j'' = \lleft(\tj)$ \textbf{iff} $x_{\sdim_{\tj}} \le \sloc_{\tj}$ 
\State %
          $\samplepartialMPblock\bigl( j'', \D,  \minsamples \bigr)$ %
\Else

\State Update $\bell_j^x \gets \min(\bell_j^x, \bx), \bu_j^x \gets \max(\bu_j^x, \bx)$ \algcomment{update extent of node $j$}
\If{$j\notin\leaf{\PT}$}  \algcomment{return if $j$ is a leaf node, else recurse down the tree}
\State \textbf{if} $x_{\sdim_j}\leq \sloc_j$ \textbf{then} $\childj=\leftj$ \textbf{else} $\childj=\rightj$
\State $\extendMPblock(\TS,\childj,\D,\minsamples)$  \algcomment{recurse on child containing $\D$}
\EndIf
\EndIf
\end{algorithmic}
\end{algorithm*}

\newcommand{\predmean}{\E[y]}
\newcommand{\predsecondmom}{\E[y^2]}
\newcommand{\predprob}{\predictive{\TS}}
\begin{algorithm*}%
\caption{$\predict\bigl(\TS,\bx\bigr)$}
\label{alg:predict}
\begin{algorithmic}[1]
\State \LineComment{Description of prediction using a Mondrian tree given by (\ref{eq:pygivenx_tree}).}
\State \LineComment{The predictive mean, predictive variance and NLPD computation are not shown, but they can be computed easily during the top-down pass using the weights $w_j$ and posterior moments $m_j, v_j$ at node $j$.} 
\State Initialize $j=\root$ and $\pnotsplityet=1$
\While{$\mathsf{True}$}
\State Set  $\Delta_j=\tt_j-\tt_{\parent(j)}$ and $\eta_j(\bx)=\sum_d\bigl(\nonnegative{x_d-u_{jd}^x}+\nonnegative{\ell_{jd}^x-x_d}\bigr)$
\State Set $p^{s}_j(\bx)=1-\exp\bigl(-\Delta_j\eta_j(\bx)\bigr)$
\If{$p^{s}_j(\bx)>0$}
\State $w_j =  \pnotsplityet  \ p^{s}_j(\bx) $
\EndIf
\If{$j\in\leaf{\PT}$}
\State $w_j  = \pnotsplityet (1-p^s_j(\bx))$
\State \Return %
\Else
\State $\pnotsplityet \gets \pnotsplityet (1-p^{s}_j(\bx))$
\State \textbf{if} $x_{\sdim_j}\leq\sloc_j$ \textbf{then} $j\gets\leftj$ \textbf{else} $j\gets\rightj$ \algcomment{recurse to the child where $\bx$ lies}
\EndIf
\EndWhile
\end{algorithmic}
\end{algorithm*}

\section{Choosing the hyperparameters}
\label{app:hyper}

In this appendix, we give more details on how we choose the hyper parameters $\btheta=\{\mu_H, \gamma_1, \gamma_2, \sigma_y^2\}$.  For simplicity, we used the same values of these hyperparameters for all the trees; it is possible to optimize these parameters for each tree independently.

We optimize the \emph{product of label marginals}, integrating out $\bmu$ for each label individually, i.e.,
\begin{align*}
q(Y|\btheta, \TS) 
= \prod_{j\in\leaf{\PT}} \prod_{n\in N(j)} \Normal(y_n|\mu_H, \phi_j-\phi_{\parent(\root)}+\sigma_y^2).
\end{align*}
Since $\tt_j=\infty$ at the leaf nodes, we have
\begin{align*}
\phi_j-\phi_{\parent(\root)} &= \mysigmoid{\tt_j}-\mysigmoid{0} \\
&= \gamma_1 (\sigma(\infty) - \sigma(0)) \\
&= \frac{\gamma_1}{2}.
\end{align*}
If the noise variance is known, $\sigma_y^2$ can be set to the appropriate value. In our case, the noise variance is unknown; hence, we parametrize $\sigma_y^2$ as $\gamma_1/K$ and set $K=\min(2000, 2N)$ to ensure that the noise variance $\sigma_y^2$ is a non-zero fraction of the total variance $\gamma_1/2+\gamma_1/K$. 

We maximize $q(Y|\btheta,\TS)$ over $\mu_H$, $\gamma_1$, and $K$, leading to 
\begin{align*}
\mu_H &= \frac{1}{N} \sum_n y_n, \nonumber\\
\gamma_1(\frac{1}{2}+\frac{1}{K}) &= \frac{1}{N} \sum_n (y_n-\mu_H)^2.
\end{align*}
Note that we could have instead performed gradient descent on the actual marginal likelihood produced as a byproduct of belief propagation.  It would be interesting to investigate this.

The likelihood $q(Y|\btheta,\TS)$ does not depend on $\gamma_2$, and so we cannot choose $\gamma_2$ by optimizing it. 
We know, however, that $\tt$ increases with $N$. 
Moreover, \citet{MF} observed that the average tree depths were 2-3 times $\log_2(N)$ in practice. 
We therefore pre-process the training data to lie in $[0,1]^D$ and set $\gamma_2=\frac{D}{20\log_2 N}$ since (i) $\tt$ increases with tree depth and the tree depth is $\O(\log_2 N)$ assuming balanced trees
and (ii) $\tt$ is inversely proportional to $D$. In Appendix~\ref{app:fast}, we describe a fast approximation which does not involve estimation of $\gamma_1, \gamma_2$. 

\section{Fast approximation to message passing and hyperparameter estimation}
\label{app:fast}

In Section~\ref{sec:bayesopt}, we suggested a fast $\O(\log n)$ approximation to exact message passing which costs $\O(n)$. Under this approximation, the Gaussian posterior at each node is approximated by a Gaussian distribution whose mean and variance are given by the  empirical mean and variance of the data points at that node. This approximation is better suited for online applications since adding a new data point involves just updating mean and variance for all the nodes along the path from root to a leaf. Another advantage of this approximation is that we only need to set the noise variance $\sigma_y^2$ and do not need to set the hyper-parameters $\{\mu_H, \gamma_1, \gamma_2\}$. 

Since our initial publication, we have learnt that this Gaussian posterior approximation is similar to a random forest modification independently proposed in \citet[\S4.3.2]{hutter2014algorithm}. In \citep{hutter2014algorithm}, each tree outputs a predictive mean and variance equal to the empirical mean and variance of the labels at the leaf node of the decision tree. However, there is an additional level of smoothing in MFs that is not present in \citep{hutter2014algorithm}. Specifically, the prediction from a Mondrian tree, described in (\ref{eq:pygivenx_tree}), is a weighted mixture of predictions from nodes along the path from the root to the leaf. Moreover, the weights account for the distance between the test point from the training data, thereby ensuring that the predictions shrink to the prior as we move farther away from the training data. %

\end{document}